Proof version. Published in the proceedings of the Intelligent Systems Conference 2026. In: K. Arai and P. Lorenz (eds.), Intelligent Systems and Applications, LNNS 2075, pp. 737-756, Springer, Cham (2027). https://doi.org/10.1007/978-3-032-32732-1_41

Check for updates

# CIRCLE: A Framework for Evaluating AI from a Real-World Lens

Reva Schwartz[1(✉)], Carina Westling[2], Morgan Briggs[3], Marzieh Fadaee[4], Isar Nejadgholi[5], Matthew Holmes[6], Fariza Rashid[7], Maya Carlyle[8], Afaf Taïk[9], Kyra Wilson[10], Peter Douglas[11], Theodora Skeadas[12], Gabriella Waters[1,13], Rumman Chowdhury[14], and Thiago Lacerda[3]

[1] Civitaas Insights, Washington, DC, USA
reva@civitaas.com
[2] Bournemouth University, Poole, UK
[3] Independent Researcher, London, UK
[4] Cohere Labs, Toronto, Canada
[5] National Research Council Canada, Ottawa, Canada
[6] Intellect Frontier, London, UK
[7] University of Sydney, Sydney, Australia
[8] National Physical Laboratory, Teddington, UK
[9] Université de Sherbrooke, Sherbrooke, Canada
[10] University of Washington, Seattle, WA, USA
[11] Principled AI, Sydney, Australia
[12] Humane Intelligence Non-profit, Katy, TX, USA
[13] Virginia State University, Petersburg, VA, USA
[14] Humane Intelligence Public Benefit Corporation, San Francisco, CA, USA

**Abstract.** This study proposes CIRCLE, a six-stage, lifecycle-based framework to bridge the "reality gap" between model-centric performance metrics and AI system outcomes in deployment. Current approaches such as MLOps frameworks and AI model benchmarks offer detailed insights into system stability and model capabilities, but they do not provide decision-makers outside the AI stack with systematic evidence of how these systems actually behave in real-world contexts or affect their organizations over time. CIRCLE operationalizes the "Validation" phase of TEVV (Test, Evaluation, Verification, and Validation) by translating priorities of stakeholders outside the stack into measurable signals. Unlike participatory design which often remains localized, or algorithmic audits which are often retrospective, CIRCLE provides a structured, prospective protocol for linking context-sensitive qualitative insights to scalable quantitative metrics. By integrating methods such as field testing, red teaming, and longitudinal studies into a coordinated pipeline, CIRCLE produces "systematic knowledge"—evidence that is comparable across sites yet sensitive to local context. This can enable governance based on materialized downstream effects rather than theoretical capabilities.

This work was conducted as part of the Forum for Real-World AI Measurement and Evaluation (FRAME) at Virginia State University's Center for Responsible AI.

## 1 Introduction

As adoption of generative AI continues to rise, policymakers, organizations, and the public are increasingly asking not just what these tools can do, but whether they can work safely and create meaningful value in the real-world [1,2]. Organizations want to identify viable AI use cases in advance and convert pilot deployments into durable successes, while the public expects that AI will work as advertised and not expose them, their families, or their communities to unsafe or harmful outcomes [2–5]. AI evaluation is the applicable discipline for answering these questions, yet the development-centric lens of the current ecosystem is unable to accommodate the diversity of purposes and settings in which these technologies are being adopted and used [6]. The lifecycle described in this study aims to address this gap by formalizing methods, tools, and processes that can produce systematic knowledge to support decision making beyond the AI development stack.

Like other forms of technology, AI can contribute to long-term and broad-scale impacts that reshape human behavior, organizational structures, and societal processes over time [7]. These higher-order effects underscore the importance of evaluating AI from a real-world and purpose-driven lens. The dominant evaluation ecosystem focuses on AI's primary effects—the immediate outputs of a model in isolated settings—rather than the secondary effects those outputs produce in everyday life. For example, "AI psychosis" and other psychological effects linked to users' over-reliance on chatbots do not result from any single output, but from cumulative interaction patterns between people and AI systems over time [8–10].

As AI is embedded into workplace operations, secondary effects can manifest as shifts in day-to-day workflows, redefinitions of job roles, and changes in how teams coordinate and make decisions [11–14]. Over longer horizons, such changes can accumulate into tertiary effects on productivity, profitability, and overall return on investment (ROI) [15,16]. AI's ROI is currently estimated using self-reported data from surveys with executives [17]. These measures are more likely to capture executive expectations, narratives, and strategic signaling than realized operational changes, and are difficult to corroborate without direct measures of deployed AI and its associated impacts [18,19]. With AI's broad-scale societal shifts largely invisible to the current evaluation ecosystem, policymakers and organizations have limited insight to inform decisions about whether and how to deploy these systems in everyday settings.

The CIRCLE lifecycle shifts the scope of evaluation to address the gap between static, model-centric benchmarks and the dynamic higher-order effects posed by the use of AI across a variety of contexts. Current evaluations often treat such contextual factors as noise to be eliminated [20–22], which can threaten core forms of measurement validity—such as ecological, construct, and consequential validity [23, 24]—and undermine the credibility of inferences about how models will behave once deployed [20, 25]. The CIRCLE lifecycle leverages context as the primary signal of system behavior to generate systematic knowledge about how AI systems perform in real-world deployments. Specifically, it delivers:

1. *An Integrated Lifecycle:* A six-stage architecture that unifies currently isolated evaluation activities—such as stakeholder elicitation (Stage 1), red teaming (Stage 3), and longitudinal monitoring (Stage 6)—into a coherent, iterative feedback loop. This brings a traceable pipeline to current ad-hoc testing paradigms where every stage generates its own documented output, feeding directly into the next stage to build a continuous record of evaluation.
2. *A Construct Operationalization Schema:* A method for translating the concepts that matter to stakeholders outside the AI stack (e.g., "over-reliance" or "cognitive offloading") into observable behavioral indicators and quantitative metrics. This process bridges the gap between qualitative socio-technical concerns and quantitative measurement—yielding a context brief that serves as the foundation for subsequent stages.
3. *Scaffolding for Assessing Higher-Order Effects:* A formalized process and scenario set that guides evaluators through structured testing of AI technologies to capture and characterize real-world consequences. It extends measurement beyond immediate model outputs (primary effects) to systematically document downstream impacts—such as shifts in workflow, decision-making authority, or long-term skill retention (secondary and tertiary effects).

Definitions for key AI systems terminology are listed in the "AI Deployment and System Terms" box below.

**AI Deployment and System Terms**

1. **AI deployment**: Phase of a project where a system is put into operation and cutover issues are resolved [26].
2. **AI model capabilities**: The tasks, functions, and behaviors an AI model can reliably perform, given specified inputs and conditions.
3. **AI stack**: A layered set of technologies, tools, frameworks, and infrastructure for building, deploying, and operating AI systems and applications.
4. **AI system**: A machine-based system that, for explicit or implicit objectives, infers, from the input it receives, how to generate outputs such as predictions, content, recommendations, or decisions that can influence physical or virtual environments. Different AI systems vary in their levels of autonomy and adaptiveness after deployment [27].
5. **Generative AI**: A category of AI that can create new content such as text, images, videos and music [28].
6. **Predictive AI**: AI systems whose primary function is to infer from input data and produce predictions or forecasts about future states, behaviors, or outcomes, typically to inform decisions or actions [27].

## 2 The CIRCLE Framework

CIRCLE—shortform for *Contextualize, Identify, Represent, Compare, Learn, and Extend*—shifts evaluation methodology to address the following key gaps in the current evaluation ecosystem:

1. **Treating real-world heterogeneity as signal instead of noise:** Model-centric benchmarking typically measures abstract capability under optimized conditions, filtering out real-world variability [22,29–32]. CIRCLE instead treats differences in how people interpret and adapt to AI as a primary signal of operational success, revealing gaps between technical accuracy metrics and actual deployment outcomes.
2. **Bringing a socio-technical frame to measurement:** Benchmarking uses script-based in silico testing, and MLOps frameworks focus on technical stability, latency, and data drift. CIRCLE extends evaluation to AI's higher-order effect by examining how interactions with AI systems in context trigger mechanisms in real workflows that produce downstream outcomes for people and organizations.
3. **Moving beyond evaluating AI "one context at a time" to capture contextual knowledge at scale and across settings:** Participatory and contextual methods generate rich local insights but often produce fragmented signals that are difficult to compare across sites. CIRCLE combines the contextual depth of participatory methods with the structured, repeatable discipline of industrial testing, allowing organizations to use shared constructs and metrics for comparing systems across settings.

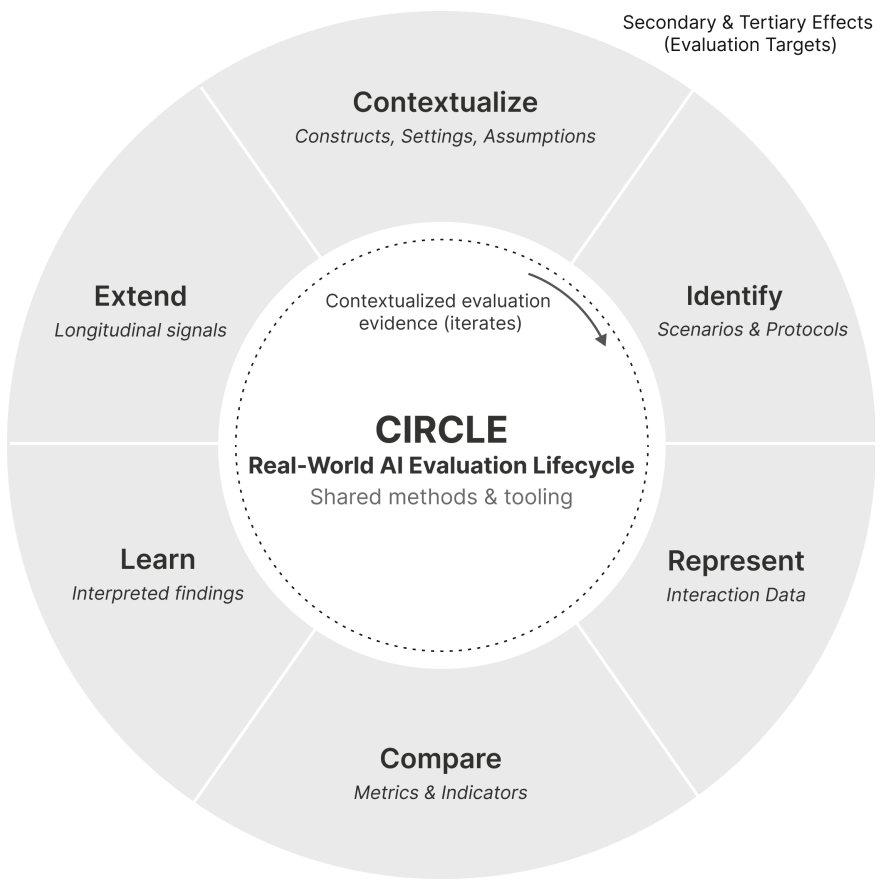

**Fig. 1.** The CIRCLE framework, illustrating its iterative structure and the primary outputs produced at each stage.

Figure 1 shows how contextualized evaluation evidence flows iteratively across the CIRCLE lifecycle stages, with a brief description of the primary output for each stage.

The lifecycle helps operationalize the "Validation" phase of the TEVV framework [33] by shifting focus from model compliance with preset criteria to what actually happens when people use AI in realistic conditions. Its main contribution is a construct-centered TEVV process that: (1) starts from stakeholder-articulated priorities about deployment—rather than available datasets, (2) systematizes those priorities into measurable constructs—rather than using convenient proxies, and (3) coordinates evaluation methods within a single, traceable pipeline that can be run in situ and at scale—rather than as one-off, small-sample or purely lab-based tests. By tying every metric back to a named construct and a stakeholder-relevant outcome, CIRCLE produces evidence that is not just technically accurate, but operationally valid for deployment-level decision-making.

This section uses an EdTech vignette to illustrate how each stage operates across the full lifecycle. Figure 2 shows how stakeholder priorities are elicited, organized into constructs, and linked to observable behaviors and higher-order effects for the EdTech vignette. In this example, stakeholders identify teacher and student over-reliance on AI chatbots as a key evaluation priority since it can lead to changes in instructional quality and learning over time. Stage-specific figures highlight key activities by emphasizing the focal stage and greying out the others in the sequence.

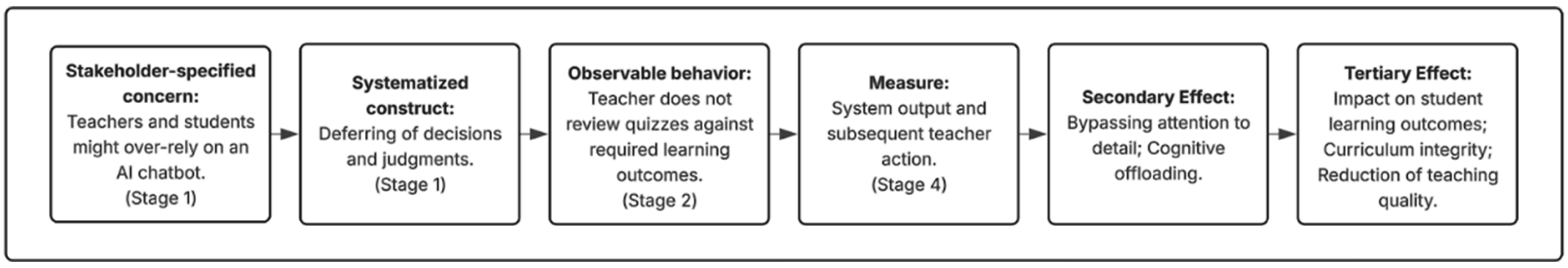

**Fig. 2.** Tracing a stakeholder-specified concern about over-reliance on an AI chatbot through the full lifecycle, from construct formation to observable behavior and longer-term outcomes in an EdTech classroom.

## 2.1 Stages of the Lifecycle

The CIRCLE lifecycle defines what matters in a setting, runs context-aware tests, interprets results for decision-makers, and feeds insights into ongoing monitoring. Each stage produces a deliverable that becomes the next stage's input, forming a traceable chain of evaluation work products. Table 1 summarizes the key concepts and guiding questions associated with each stage.

CIRCLE differentiates itself by centering people who test AI systems as situated experts rather than one-off labelers of model outputs. It asks them how system behavior aligns with their local goals, constraints, and norms, then pairs their accounts with interaction transcripts to surface real-world usage patterns tied to stakeholder priorities. These human-centered collections can run alongside large-scale automated tests so that findings generalize beyond a narrow test group or evaluation mode, with each lifecycle stage linking stakeholder-defined concepts to concrete methods and evidence.

**Table 1.** Concepts and questions at each stage of the CIRCLE framework

| Framework element and lifecycle stage | Designed to answer: |
|---|---|
| *Contextualize*<br>**Context specification stage**<br>Sets forth key concepts of interest in the deployment context from the viewpoint of those outside the AI stack<br>**Output deliverable:** Context brief | What specific human, system, or interaction properties do the stakeholders expect in this setting? |
| *Identify*<br>**Evaluation design and planning stage**<br>Translates stakeholder-defined concepts of interest into test-ready methods and data-collection protocols<br>**Output deliverable:** Evaluation Design Plan | What forms of evidence need to be produced in order to judge whether system outcomes can match stakeholder expectations in this setting?<br>How will that evidence be elicited, captured, and tracked over the course of the evaluation?<br>What technical and organizational infrastructure is required to run this evaluation? |
| *Represent*<br>**Evaluation execution stage**<br>Executes testing to collect representative data as specified in the evaluation design and planning stage<br>**Output deliverable:** Evaluation Execution Plan | How will the test subjects interact with systems to produce evidence laid out in the evaluation plan?<br>What populations and use patterns are required for the evaluation scenarios?<br>What infrastructure processes and resources will be used during the evaluation period? |
| *Compare*<br>**Analysis stage**<br>Synthesizes cross-cutting evaluation evidence to tie outcomes to concepts of interest<br>**Output deliverable:** Findings Synthesis Report | Which narratives or tasks can surface the systematized constructs in a given setting, and how can those signals be captured for streamlined analysis?<br>What kinds of observations are needed to tie materialized evaluation outcomes to the constructs of interest? |
| *Learn*<br>**Insights stage**<br>Enables stakeholders outside the AI stack to make sense of, prioritize, and act on evaluation outcomes<br>**Output deliverable:** Stakeholder Insights Brief | Which stakeholders should be considered when writing and disseminating insights?<br>What type of publications or platforms will optimize dissemination and engagement with implementation of the insights? |
| *Extend*<br>**Continuous monitoring stage**<br>Tracks shifts in AI deployment from the perspective of stakeholders beyond the AI stack<br>**Output deliverable:** Continuous Monitoring Plan | Are post-evaluation controls operating as intended?<br>Which stakeholders and/or domain experts should be involved to identify major shifts in context?<br>What resources are required to conduct ongoing monitoring and reporting? |

Patterned on Chouldechova et al.'s shared standard for valid measurement of generative AI systems [34], the CIRCLE lifecycle shifts their core categories from Concepts to Concepts *and* Questions; Instance to Processes; Population to Who and Where; and Amounts to Outcomes. It centers practical context, materialized impacts, and stakeholder input, broadening who defines evaluation questions, their scope, and what counts as evidence. Full descriptive details for each lifecycle stage appear in Table 2 in the Appendix.

Real-world evaluation reaches beyond the technical stack into the settings where people decide whether and how to use AI. This broader remit requires different organizational capabilities and resourcing than conventional model testing. For example, organizations must be able to support stakeholder engagement, construct articulation, instrument design, and quasi-experimental testing with large and sufficiently diverse populations across various settings. The ability to conduct and sustain these activities over time may be out of reach for many individual teams acting alone, underscoring the need for shared methods, tools, and support structures that lower the barrier to participation in robust, context-sensitive evaluation.

The "AI Evaluation and Testing Terms" box below lists definitions for core evaluation terms found in this study.

**AI Evaluation and Testing Terms**

- **Benchmark/benchmarking**: Procedure, problem, or test that can be used to compare systems or components to each other or to a standard [26].
- **In silico testing**: Testing or experimentation carried out entirely on a computer, using computational models and simulations.
- **Real-world AI evaluation**: The process of accounting for what materializes when people use AI systems in practical, everyday contexts.
- **Testing, evaluation, validation, verification (TEVV)**: A framework for assessing and incorporating methods and metrics to determine that a technology or system satisfactorily meets its design specification and requirements and that it is sufficient for its intended use [33].

**Stage 1: Context Specification.**
AI's higher-order effects reflect how users, communities, organizations, and societies evolve as these technologies are introduced, adopted, and normalized [7,35–37]. Disentangling these effects and their contributing factors requires "contextual awareness", a process that helps stakeholders identify and make sense of the situational factors that shape AI's role, purpose, and impacts in their own settings [38,39].

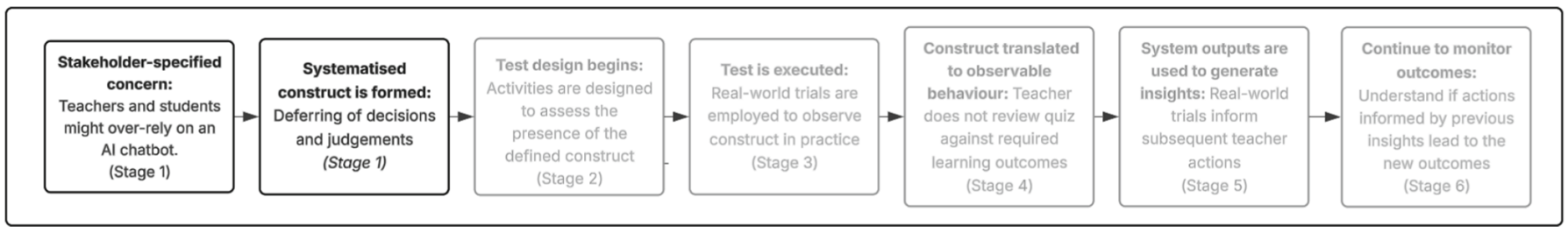

**Fig. 3.** Stage 1 of the lifecycle in the EdTech example: eliciting and systematizing stakeholder concerns about over-reliance on an AI chatbot, with later stages shown in grey.

Since situational factors are experienced differently across roles and communities, the lifecycle's first stage builds contextual awareness by eliciting what matters most to stakeholders in a given setting [22,40–44]. This grounding step provides the basis for interpreting results in later stages and for judging whether observed outcomes truly reflect the concerns, priorities, and constraints of stakeholders closest to the deployment context. Elicitation can happen through in-person workshops, interviews, and process mapping, or virtually via asynchronous exercises, offering flexibility around time and resource constraints. These techniques surface both explicit knowledge (policies, procedures, formal requirements) and tacit knowledge (experiential insights, practical judgment, uncodified understanding) needed to grasp the deployment context and how organizations actually operate [45].

Stakeholder-elicited priorities can be augmented by automated methods, such as LLM-based review of documented curricula, policies, and community materials, to surface considerations that stakeholders may overlook [46–48]. Combining human-centered elicitation with automated techniques offers a more comprehensive picture of both tacit and explicit knowledge, aligning with work that frames AI and humans as complementary partners in knowledge work. This approach also makes it feasible to extend contextual analysis across many settings that match the prospective use case [49].

In the education example, school leaders, educators, and communities begin by specifying their priorities, concerns, and expectations for the classroom chatbot. Evaluators then situate these priorities within supporting information about the deployment context—the system's purpose, operational constraints, regulatory and ethical boundaries, and the stakes for those affected. This collected information is then synthesized into an evaluation construct: a clear, named idea representing the specific risk, behavior, or value that stakeholders need evidence about to support their deployment decisions. For example, identification of an "over-reliance" construct might cover cognitive offloading and preserving human judgment in the classroom. This can be articulated as "teachers and students might defer their own decisions and judgments to an educational chatbot". Figure 3 shows how Stage 1 elicits and systematizes these stakeholder priorities, producing a "context brief" that feeds into Stage 2.

**Stage 2: Evaluation Design and Planning.**
Stage 2 formalizes the construct laid out in the context brief so it is specific enough to guide observation and measurement - for example “changes in content review and deferring of decisions and judgments”. Evaluators then design evaluation methods and scenarios that make the construct measurable and observable within the relevant context (e.g., the classroom setting). This foregrounding of context-specific factors contrasts with model-centric evaluations, which typically use static datasets, canned prompts, and tightly controlled test environments. Relevant contextual detail helps stakeholders beyond the stack make sense of evaluation results in their own settings to support informed deployment decisions. Figure 4 show how Stage 2 activities operationalize these design choices for the EdTech example.

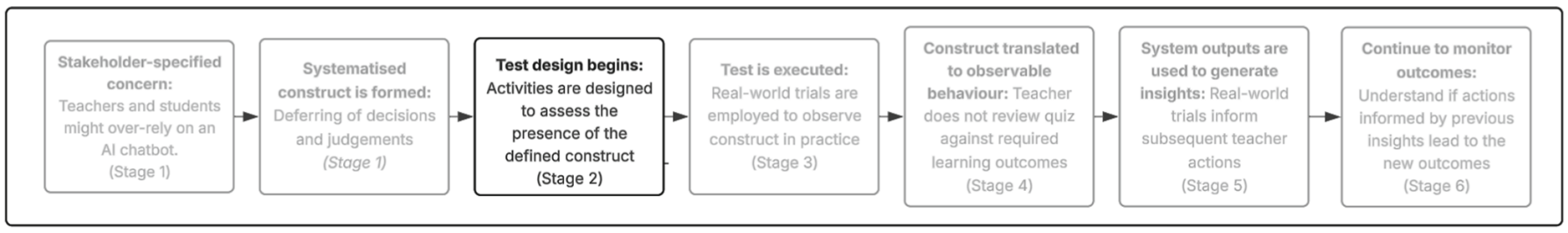

**Fig. 4.** Stage 2 of the lifecycle in the EdTech example: designing test activities to assess the presence of the defined construct.

Figure 5 shows a continuum of evaluation methods that can be selected and combined to deploy the Stage 2 test scenarios, highlighting tradeoffs between pre-scripted, in-silico tests and context-rich, real-world methods, and how those choices shape what forms of evidence can be observed.

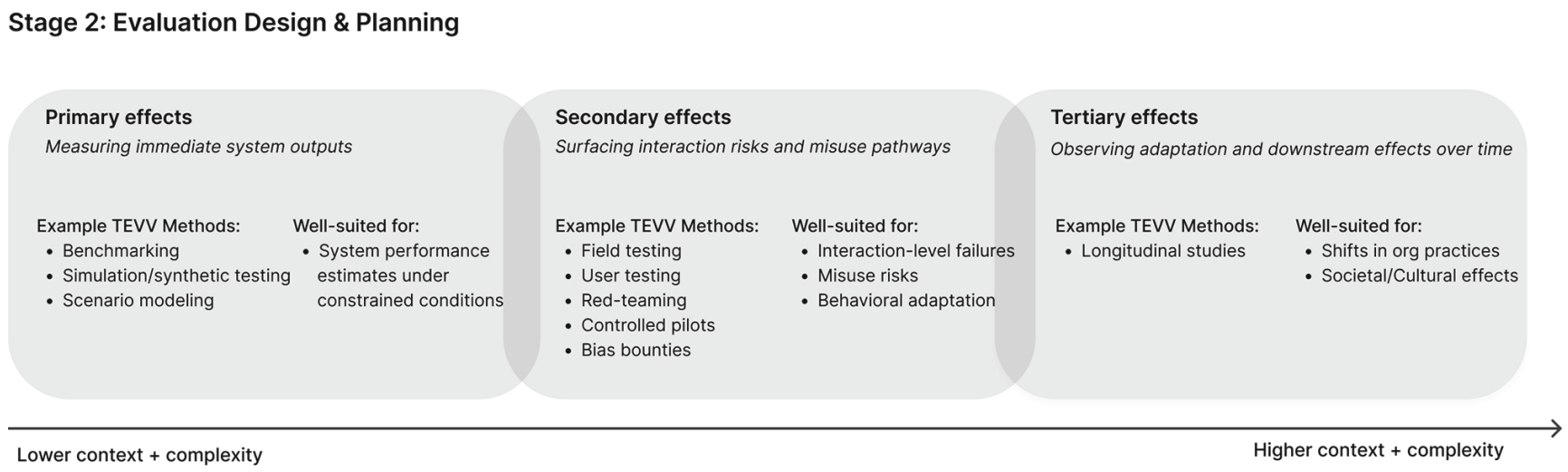

**Fig. 5.** Evaluation methods as design choices in Stage 2 of the CIRCLE lifecycle. Methods range from pre-scripted, in-silico tests to context-rich, real-world studies, shaping what forms of evidence and downstream effects can be observed. Regions illustrate classes of methods that may be selected and combined during this stage.

At the primary end of the continuum is AI benchmarking, which focuses on whether models produce “correct” or expected outputs—yielding metrics such as accuracy, efficiency, and energy use. By examining whether people can successfully navigate AI interfaces to complete tasks, methods such as usability studies begin to surface AI's secondary effects—revealing user friction, confusion, or drop-off points that simple benchmarks cannot detect.

Context-sensitive methods such as red-teaming, bias bounties, and field tests evaluate system behavior during live interactions with people under realistic or adversarial conditions [50,51]. At the tertiary end of the continuum, longitudinal studies can be combined with other real-world signals like incident tracking to monitor cumulative downstream effects that no single snapshot can capture, such as how innovation benefits for organizations, sectors, and the broader economy evolve over time.

The "Contextual Measurement Concepts" box below lists definitions for terms related to construct development.

**Contextual Measurement Concepts**

- **Construct**: An abstract, latent concept or theoretical variable, not directly observable, that is defined for scientific purposes and measured indirectly through multiple observable indicators or items.
- **Construct systematization**: The process of clarifying and organizing a concept by specifying its meaning, dimensions, and relationships to other concepts [52].
- **Construct operationalization**: Linking a systematized concept to appropriate indicators and scores in a coherent measurement framework [52].
- **Context**: The parameters in which interrelated factors, purposes, and circumstances may shape individual and collective perceptions, interpretations, and expectations about the functionality and impacts of AI technology and resulting actions [53].

**Stage 3: Evaluation Execution.** Testing is carried out in Stage 3 through real-world trials of AI systems with large, diverse participant groups so that findings can generalize across user categories and contexts [54] (illustrated in Fig. 6). This stage also defines participant sampling based on evaluation requirements and in line with human subjects research protocols, participant incentives, and expected attrition. Since AI's higher-order effects can impact both users and non-users, evaluators can use stratified sampling and oversampling to recruit people who do not use AI so their behaviors are represented. Non-user responses can then be up-weighted in analysis to reduce adopter-centric bias and better represent groups currently underserved by chat models [55,56]. For example, in EdTech this might require enrolling teachers from different schools and with different seniority levels and experience with AI [57].

Participants receive training before testing so they can use the system safely and understand their role and the system's intended use. They are provided informed consent and a code of conduct covering ethical considerations, data collection, acceptable use, and how to report or escalate harms or unexpected behavior [58]. Codes of conduct should also describe core system functional-ity, known failure modes, and key limitations in plain language. Depending on the evaluation design, participants may also be trained to conduct adversarial testing.

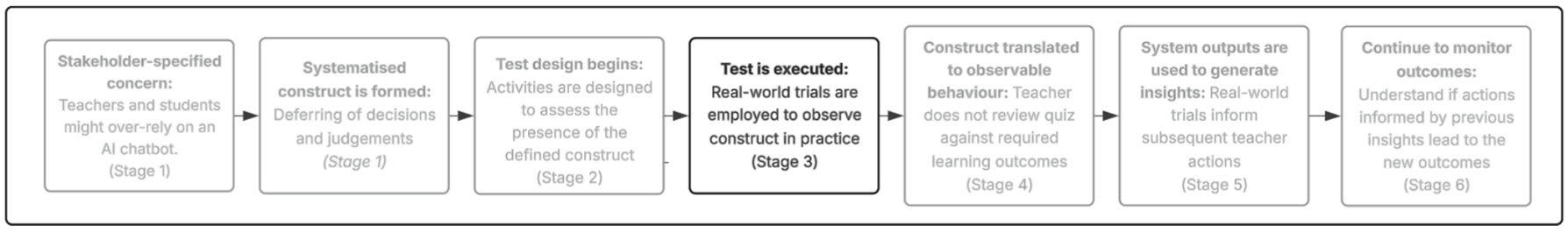

**Fig. 6.** Stage 3 of the lifecycle in the EdTech example: executing real-world trials to observe the construct in practice.

Fully automated tests can run alongside participant trials using the same test harness, enabling directly comparable evaluation outcomes and providing a more comprehensive view of the deployment environment. Decisions across the lifecycle should be tracked and documented so evaluation outcomes can be interpreted in light of testing mode, conditions, and participant category. Hypotheses, metrics, sampling plans, and protocols defined in the design stage typically require adjustment due to factors such as recruitment shortfalls or technical difficulties that lead to unexpected usage patterns or missing data. Documenting these deviations is essential for sound analysis in subsequent stages and for deciding whether to repeat, adapt, or add experiments.

**Stage 4: Analysis.** Stage 4 supports scenario design by creating the scaffolding needed to make constructs of interest observable and measurable [59,60]. For example, systematically observing how often a teacher reviews AI-generated quizzes for required learning outcomes makes the over-reliance construct concrete and analyzable. This behavior can be tracked through student quiz results, chatbot transcripts, and records of any follow-on actions—yielding insights into higher-order effects within and across classrooms. Proxy scenarios can be used to assess hard-to-observe higher-order effects without directly exposing participants to harmful content or experiences.

Scenarios anchor the development of other analysis tools, including scoring rubrics, scoring criteria, and data markup methods. Scoring criteria are designed to connect the elicited evaluation behaviors to what actually materializes in context. Rubrics are used to consistently map raw observations onto outcome metrics—all of which are tied to stakeholder-articulated priorities. Annotation and markup of output data can be fully automated, fully manual, or participant-driven, but must use universally understood definitions for the observed behaviors to be sure that everyone applies and interprets the definitions consistently throughout the markup process. Figure 7 highlights Stage 4, translation of constructs into observable behaviors and indicators.

Analysis can draw on qualitative results from classroom trials and large-scale model runs so each informs the other in context rather than standing alone. Observed behaviors and coded patterns from contextualized student-chatbot transcripts and other data are then compared to automated metrics from scenario-matched test suites. Because both evaluation modes draw on the same constructs, their combined measures can connect specific deployment conditions, model behavior under those conditions, and the secondary effects that emerge at scale. This context-rich information, in turn, supports broader inferences about tertiary effects that link concrete classroom practices to longer-term shifts in learning outcomes, curricula, and teaching effectiveness.

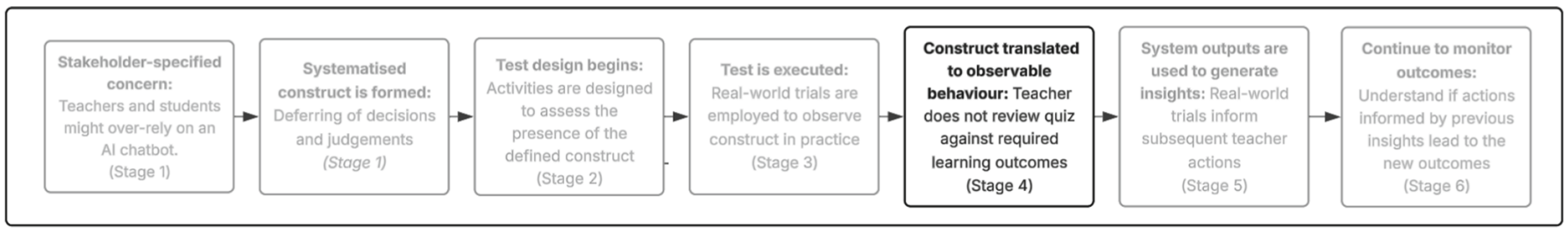

**Fig. 7.** Stage 4 of the lifecycle in the EdTech example: translating the construct into observable teacher behaviors and classroom signals.

Definitions for key testing concepts described in Stage 4 are listed in the "Testing Scenario Concepts" box below.

## Testing Scenario Concepts

- **Evaluation scenario**: A high-level description of a specific situation or sequence of conditions under which a system is to be tested, including the initial state, inputs, triggering events, and expected behavior or outcomes.
- **Proxy scenario**: A test scenario that does not directly reproduce the real-world context of interest but is designed to stand in for it, using more tractable or safer conditions while preserving key features believed to be relevant for evaluating system behavior or risk.

**Stage 5: Insights.** Deriving evaluation insights involves putting analysis findings back into context and extrapolating them to likely future conditions. Depending on how the participant and automated runs are designed, their combination can support longitudinal projections and examination of tertiary effects. Bringing participants and stakeholders into guided discussions of local test results helps re-contextualize the findings and translate them into insights that meet their interpretive needs.

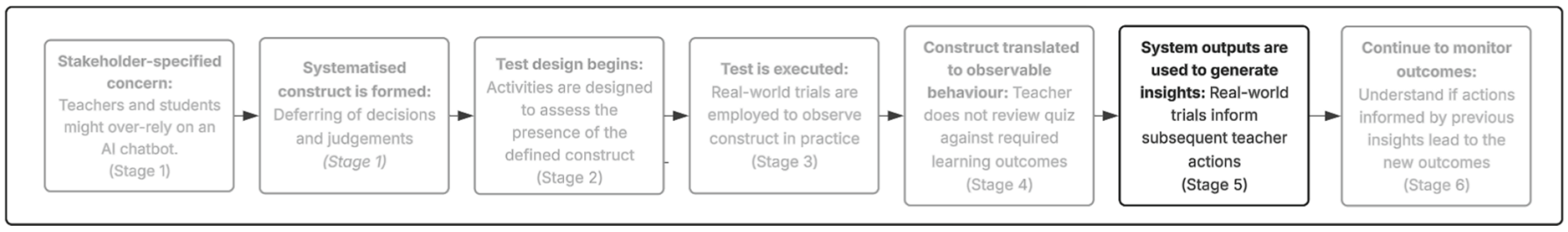

**Fig. 8.** Stage 5 of the lifecycle in the EdTech example: using observed system outputs to generate insights that inform subsequent teacher actions.

Interpretive methods are context-sensitive and should consider audiences, reach, and impact and can make use of the concepts identified by stakeholders in Stage 1. Findings should be presented in ways that match the needs and communication styles of the intended audiences, relying on plain language for maximum uptake beyond technical audiences. Figure 8 shows how EdTech evaluation outputs can be used in Stage 5 to generate insights and inform the stakeholder audience. Stakeholders can support selection of appropriate channels—such as industry periodicals, newsletters, internal reports, conferences, journals, news media, and community networks—and shape tone and format, so materials are easy to find, understand, and act on.

**Stage 6: Continuous Monitoring.** Stage 6 establishes continuous, context-aware monitoring of deployed AI systems to detect performance drift, emerging risks, and changes in real-world usage patterns as deployment conditions evolve [61,62]. Monitoring targets are anchored in the constructs and risk hypotheses identified during context specification, while also incorporating exploratory signals to surface unanticipated shifts in system behavior or use [63]. In the education example, monitoring tracks how chatbots are actually used in evolving classroom environments so that over-reliance, drifting learning outcomes, or misalignment with institutional or ethical guidelines can be detected and addressed early [64]. Figure 9 highlights how Stage 6 monitoring operates across multiple organizational and social layers, each capturing distinct mechanisms through which AI systems shape outcomes over time. The educational targets at the district level would focus on aggregate patterns such as policy compliance, consistency of implementation across schools, and equity impacts between schools.

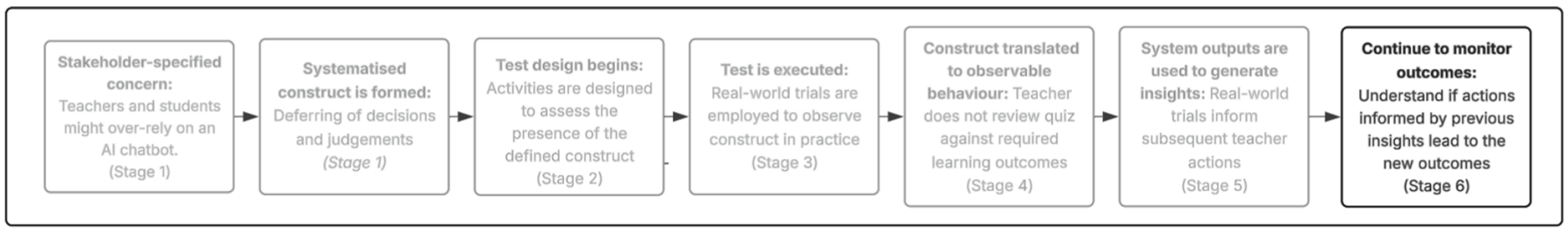

**Fig. 9.** Stage 6 of the lifecycle in the EdTech example: monitoring outcomes over time to see how actions informed by previous insights affect student learning and reliance.

At the school level, monitoring examines institutional conditions such as leadership priorities, technical infrastructure, and staff training that mediate whether AI tools are adopted, adapted, or bypassed in practice. At the classroom level, monitoring foregrounds pedagogical context, including changes in curriculum, teaching strategies, and classroom culture or demographics that influence how teachers and students interact with the system. Finally, at the individual student and educator level, monitoring assesses changes in motivation, task engagement, digital literacy, and trust calibration with respect to AI-supported decision-making.

Practically, effective monitoring in education is a hybrid process [65]. Automated systems can collect signals such as usage patterns and learning outcomes.

Human and institutional review is required to interpret these signals in relation to local goals, constraints, and unintended consequences, and to determine whether corrective action, redesign, or policy updates are warranted. The goal is a recurring cycle—observe, interpret, diagnose, act—that supports continuous improvement without collapsing monitoring into continuous surveillance of individuals or real-time behavioral control [66]. Accordingly, monitoring follows principles of data minimization, proportionality, and purpose limitation [67,68], ensuring that oversight remains legitimate, auditable, and aligned with institutional governance and privacy obligations. Findings from continuous monitoring feed back into renewed context specification and evaluation design, allowing constructs, scenarios, and metrics to be revised as deployment conditions and stakeholder priorities change.

## 3 Discussion and Future Work

CIRCLE treats AI evaluation as a construct-centered, lifecycle process that links stakeholder priorities to contextualized scenarios, mixed-methods testing, and ongoing monitoring. Instead of treating benchmarks, red teaming, and field trials as separate activities, it integrates them into one pipeline that ties every metric to a named construct, an observable behavior, and a stakeholder-relevant outcome. This makes it possible to generate evidence about the concepts that matter to people outside the AI stack so they can make informed decisions about whether and how systems should be adopted and governed in their own settings.

Implementing the CIRCLE lifecycle introduces trade-offs regarding cost and complexity. Unlike automated benchmarks, collecting context-aware information requires organizational maturity, interdisciplinary expertise, and time-intensive human-subjects protocols, and is inherently slower and more resource-intensive than running in silico test scripts. These investments, however, strengthen measurement validity and enable organizations to surface higher-order, context-specific insights about downstream and tertiary effects that directly inform deployment, governance, and decommissioning decisions.

CIRCLE enhances *ecological validity* by structuring evaluations in settings that approximate everyday life rather than in tightly controlled, artificial test environments. It strengthens *construct validity* by explicitly eliciting stakeholder concerns in Stage 1, naming them as constructs in Stages 2 and 4, then designing scenarios, coding schemes, and metrics that instantiate them for evaluation. *Criterion validity* is supported by aligning evaluation outcome measures with real-world outcomes of interest, enabling more reliable predictors of deployment performance. *Consequential validity* is improved by systematically tracking benefits and harms through real-world interactions across the lifecycle, rather than inferring them only from theory or in silico testing results. Finally, *internal validity* is strengthened through deliberate design choices in Stage 2, such as comparison groups and quasi-experimental designs, and through careful documentation of protocols and deviations during execution, which help minimize alternative explanations and experimental confounds.

Future efforts will need to reduce the friction of capturing contextual detail at scale, for example by developing validated user simulators and other tooling grounded in empirical lifecycle data. Adoption of context-sensitive insights about higher-order effects will also require shifts in institutional incentives so that deployment-level evidence is valued alongside model documentation. Investing in measurement infrastructure, including shared libraries of operationalized constructs, can help make these evaluations more interoperable across settings.

## 4 Conclusion

This study presented CIRCLE as a construct-centered lifecycle framework for evaluating AI systems from a real-world perspective to address the gap between static, model-centric benchmarks and dynamic deployment outcomes. The six stages coordinate activities such as stakeholder elicitation, evaluation design, real-world testing, analysis, and monitoring into a traceable pipeline that produces systematic knowledge about how AI systems actually function in practice. This enables decision-makers beyond the AI development stack to ground adoption and governance decisions in evidence about materialized effects, not just theoretical capabilities.

**Declaration on Generative AI.** The authors used large language model tools to convert some of the references to BibTeX format, to help structure the order of concepts in the introduction, and to correct LaTeX code for formatting tables and call-out boxes. All AI-generated text and code were fully reviewed, edited, and verified by the authors prior to submission.

## Appendix

See Table 2.

**Table 2.** CIRCLE framework

| Concepts and questions | Processes | Who and where | Outcomes |
|---|---|---|---|
| *Context specification* | | | |
| What specific human, system, or interaction properties do the stakeholders expect in this setting? | Elicit and systematize background concepts associated with stakeholder claims; document assumptions, constraints, and success criteria | Stakeholders implementing the system; stakeholders who may be affected; domain experts; context specification team; evaluation design team; analysis team; continuous monitoring team; real-world setting(s) where the system will be used | Context brief listing clearly specified concepts to be evaluated (systematized constructs), relevant populations, and target parameters grounded in the implementation setting |
| *Evaluation design and planning* | | | |
| What forms of evidence are required to judge whether system outcomes match stakeholder expectations in context? How will evidence be elicited, captured, and tracked? What technical and organizational infrastructure is required? | Infrastructure plan; sampling design; evaluation design strategy; construct operationalization plan including evaluation scenarios and task protocols; subject recruitment requirements; test protocol plan | Context specification team; evaluation design team; evaluation execution team; analysis team; IRB/ethics, legal, and privacy experts | Evaluation design plan including study design; final evaluation scenarios and test protocols; sampling plan; metrics plan; infrastructure requirements; annotation and markup guidelines; checklist of anticipated positive and negative impacts |
| *Evaluation execution* | | | |
| How will test subjects interact with systems to produce evidence? What populations and use patterns are required? What infrastructure resources are used during evaluation? | Recruit, enroll, and train participants; run tasks; collect logs per protocol; monitor and quality-control protocol deviations and data quality; manage evaluation environments; ensure data protection, compliance, and governance | Evaluation design team; analysis team; continuous monitoring team; compliance experts; test subjects; sandbox environments; field sites | Recruitment plans; training materials; raw evaluation outputs; transcripts; metadata for analysis |
| *Analysis* | | | |
| Which narratives or tasks surface systematized constructs? How are signals captured for analysis? What observations link outcomes to constructs? | Analysis design and planning; scoring design; mapping tasks to constructs; mapping anticipated outcomes to scoring; data markup plan; metrics design and execution; annotation training | Domain experts; evaluation design team; context specification team; test execution team; insights team; annotation and markup personnel | Finalized metrics and analysis techniques; data markup schema; scoring tools and guidance; analysis outputs including estimates with intervals and distributional breakdowns aligned with stakeholder claims |
| *Insights and reporting* | | | |
| Which stakeholders should be considered when disseminating insights? What platforms optimize engagement and implementation? | Composition of articles for stakeholders, industry, and research audiences; networking with stakeholder communities; relationship building with policy and decision makers; engagement with appropriate dissemination platforms | Domain experts; implementation stakeholders; stakeholder communities and interest groups; evaluation design team; analysis team; policy and decision makers; conferences; editors and journals | Reports, case studies, and instantiated evaluation processes; published articles; conference papers; white papers; extended stakeholder networks and advisory groups supporting policy and decision making |

(*continued*)

**Table 2.** (*continued*)

| Concepts and questions | Processes | Who and where | Outcomes |
|---|---|---|---|
| *Continuous monitoring* | | | |
| Are post-evaluation controls operating as intended? Who identifies shifts in context? What resources are required for ongoing monitoring? | Data gathering; monitoring thresholds; cadence of monitoring activities; horizon scanning; ongoing information exchange; development of adjustment plans | Domain experts; context specification team; analysis team; evaluation design team; relevant stakeholders; advisory groups; decision and policy makers | Periodic insight and impact reports; updated research outputs; refined processes, methods, and metrics; best-practice guidance; longitudinal qualitative and quantitative insight into real-world AI impacts across contexts |